# Uncertainty and Incompleteness: Breaking the Symmetry of Defeasible Reasoning *

Piero P. Bonissone     David A. Cyrluk     James W. Goodwin[†]     Jonathan Stillman

Artificial Intelligence Program
General Electric Corporate Research and Development
Schenectady, New York 12301
Arpanet: bonissone@crd.ge.com  cyrluk@crd.ge.com  stillman@crd.ge.com

## Abstract

Two major difficulties in using default logics are their intractability and the problem of selecting among multiple extensions. We propose an approach to these problems based on integrating nonmonotonic reasoning with plausible reasoning based on triangular norms. A previously proposed system for reasoning with uncertainty (RUM) performs uncertain monotonic inferences on an acyclic graph. We have extended RUM to allow nonmonotonic inferences and cycles within nonmonotonic rules. By restricting the size and complexity of the nonmonotonic cycles we can still perform efficient inferences. Uncertainty measures provide a basis for deciding between multiple defaults. Different algorithms and heuristics for finding the optimal defaults are discussed.

## 1  Introduction

### 1.1  Motivation

The management of uncertain information in first generation expert systems, when addressed at all, has largely been left to *ad hoc* methods. This has been effective only because operational expert systems normally assume that knowledge is complete, precise, and unvarying. This fundamental assumption is a principal source of the limitation of many diagnostic systems to single fault diagnoses, and the limitation of classification systems to time-invariant phenomena.

The management of incomplete information has also lacked a clear focus, as some researchers have attempted to find its solution by defining new nonmonotonic logics, by augmenting classical logic with default rules of inference, by searching for minimal models via functional optimization, or by concentrating only on the instruments, i.e. TMSs, rather than the theory required to handle this problem.

In the past, a subset of the authors have contributed to the development of individual theories for reasoning with uncertainty and incompleteness. Bonissone has proposed RUM, a system for reasoning with uncertainty whose underlying theory is anchored on the semantics of many-valued logics [Bon87]. This system provides a representation layer to capture structural and numerical information about the uncertainty, an inference layer to provide a selection of truth-functional triangular-norm based calculi [Bon87], and a control layer to focus the reasoning on subsets of the KB, to (procedurally) resolve ignorance and conflict, and to maintain the integrity of the inference base via a belief revision system. RUM, however, does not provide any declarative representation to handle incomplete information.

Goodwin [Goo87] and Brown [BGB87] have provided such a representation by developing theories based on nonmonotonic dependency networks and algebraic equations over boolean lattices, respectively. These approaches, however, have neglected the aspect of uncertain information.

Another motivation is the existence of a new class of problems, referred to as *dynamic classification problems* [BW88], which cannot be prop-

*This work was partially supported by the Defense Advanced Research Projects Agency (DARPA) under USAF/Rome Air Development Center contract F30602-85-C-0033. Views and conclusions contained in this paper are those of the authors and should not be interpreted as representing the official opinion or policy of DARPA or the U.S. Government.

[†]Currently with Knowledge Analysis, Belmont, Massachusetts.

erly addressed without an integration of the theories for reasoning with uncertainty and incompleteness. Preliminary work in this integration have been reported by D'Ambrosio (integrating assumptions and probabilistic reasoning) [DAm88].

## 1.2 Uncertainty

The existing approaches to representing uncertainty can be subdivided into two basic categories according to their *quantitative* or *qualitative* characterizations of uncertainty. (See references [Bon87a,Pea88] for a survey of approaches to reasoning with uncertainty.)

There are two distinct quantitative approaches, which differ in the semantics of their numerical representation. One is based on *probability* theory, the other on *many-valued logics*.

Some of the more traditional techniques found among the approaches derived from probability are based on *single-valued* representations. These techniques include Bayes Rule [Pea88,Pea85], Modified Bayes Rule [DHN76] and Confirmation Theory [SB75]. A more recent trend among the probabilistic approaches is represented by approaches based on *interval-valued* representations such as Dempster-Shafer Theory [Dem67,Sha76], Evidential Reasoning [LGS86], and Probability Bounds, i.e., consistency and plausibility (see [Qui83]).

Among the approaches anchored on many-valued logics, the most notable are based on a *fuzzy-valued* representation of uncertainty. These include Necessity and Possibility Theory [Zad78], the Linguistic Variable Approach [Zad79], and the Triangular-norm based approach [Bon87,BGD87,Bon89].

With numerical representations, it is possible to define a *calculus* that provides a mechanism for propagating uncertainty through the reasoning process. Similarly, the use of aggregation operators provides summaries which can then be ranked to perform rational decisions.

Models based on *qualitative* approaches, on the other hand, are usually designed to handle the aspect of uncertainty derived from the *incompleteness* of the information, such as Reasoned Assumptions [Doy83], and Default Reasoning [Rei80]. With a few exceptions, they are generally inadequate to handle the case of *imprecise* information, as they lack any measure to quantify confidence levels [Doy83]. A few approaches in this group have addressed the representation of uncertainty, using either a *formal* representation, such as Knowledge and Belief [HM85], or a *heuristic* representation, such as the Theory of Endorsements [Coh85].

The formal approach has a corresponding (modal) logic theory that determines the mechanism by which inferences (theorems) can be *proven* or *believed* to be true. The heuristic approach has a set of context-dependent rules to define the way by which frame-like structures (endorsements) can be combined, added or removed.

In our work we use a quantitative representation of uncertainty based on the semantics of many-valued logics (T-norm operators). This is integrated with a formal qualitative approach to reasoning with incompleteness.

## 1.3 Incompleteness

Traditional predicate logics, developed for reasoning about mathematics, are inadequate as a formal framework for reasoning with incomplete information in that they are inherently *monotonic*: anything that one can derive from a set of formulae can also be derived from every superset of those formulae. It is argued that people simply don't reason this way: we are constantly making assumptions about the world and revising those assumptions as we obtain more information (see [Min75] or [McC77], for instance). Many researchers have proposed modifications of traditional logic to model the ability to revise conclusions in the presence of additional information (see, for instance, [Moo83,McC86]). Such logics are called *nonmonotonic*. Informally, the common idea in all these approaches is that one may want to be able to "jump to conclusions" which might have to be retracted as new information about the world becomes available. While a detailed discussion of nonmonotonic logics is outside the scope of this paper, we present a short overview below. (See references [BB86,Rei88,Eth88] for surveys of approaches to reasoning with incompleteness. Additionally, a number of the most important papers in the field have been collected in [Gin87].)

The most prominent of the approaches to nonmonotonic reasoning seem to fall into three broad categories: those based on *consistency*, those based on *minimization*, and those based on *epistemology*. The consistency-based approaches include McDermott and Doyle's *nonmonotonic logic (NML)* [MD80], and Reiter's *default logic* [Rei80]. Basically, $NML$ consists of first-order logic together with a modal operator $M$, where $Mf$ is intended to correspond to "$f$ is consistent." The theorems of such a set of formulae are defined to be the intersection of the fixed points of an operator $NM$. These fixed points consist of the logical closure of the original theory together with some maximally consistent set

35

of the given formulae of the form $M f$. Several deficiencies in this approach have been identified (e.g., a theory may have the fixed point $\{Mf, \neg f\}$), and an attempt to rectify these was made in [McD82]. The changes suggested by McDermott also suffer from several inherent weaknesses; these are discussed in [Moo83] and in [Eth88].

Reiter [Rei80] developed an extension of first order logic which allows the specification of default rules. These differ from *NML* in that these rules are treated as rules of inference rather than as statements within the language. Default rules are used to model decisions made in prototypical situations when specific information is unavailable. For a detailed discussion of Reiter's default logic the interested reader is referred to [Rei80]. In this section we will simply review some of the immediately pertinent ideas. A *default theory* is a pair $(D, W)$, where $W$ is a set of closed well-formed formulae (wffs) in a first order language and $D$ is a set of *defaults*. A *default* consists of a triple $< \alpha, \beta, \gamma >$: $\alpha$ is a formula called the *prerequisite*, $\beta$ is a set of formulae called the *justifications*, and $\gamma$ is a formula called the *conclusion*. Informally, a default denotes the statement "if the *prerequisite* is true, and the *justifications* are consistent with what is believed, then one may infer the *conclusion*." Defaults are written

$$\frac{\alpha : \beta}{\gamma}$$

If the conclusion of a default occurs in the justifications, the default is said to be *semi-normal*; if the conclusion is identical to the justifications the rule is said to be *normal*. A default is *closed* if it does not have any free occurrences of variables, and a default theory is *closed* if all of its rules are closed.

The maximally consistent sets that can follow from a default theory are called *extensions*. An extension can be thought of informally as one way of "filling in the gaps about the world." Formally, an extension $E$ of a closed set of wffs $T$ is defined as the fixed point of an operator $\Gamma$, where $\Gamma(T)$ is the smallest set satisfying:

$W \subseteq \Gamma(T)$,

$\Gamma(T)$ is deductively closed,

for each default $d \in D$, if the *prerequisite* is in $\Gamma(T)$, and the negations of the *justifications* are *not* in $T$, then the *conclusion* is in $\Gamma(T)$.

Since the operator $\Gamma$ is not necessarily monotonic, a default theory may not have any extensions. Normal default theories do not suffer from this, however (see [Rei80]), and always have at least one extension. One popular attempt to develop a computationally tractable approach to default logic is based on *inheritance networks with exceptions.* A thorough presentation of this can be found in [Tou86]. The second class of approaches to nonmonotonic reasoning involves reasoning about the knowledge or beliefs of an agent. Responding to the weaknesses of *NML*, Moore [Moo83] developed an epistemology-based logic called *autoepistemic logic.* Moore's logic differs from *NML* in two basic ways: the fixed points cannot contain both $L f$ and $\neg f$, and each fixed point is considered to correspond to one way that the world might be, in contrast to *NML*, where a proposition is believed only if it is present in all fixed points. Although Moore's original formulation was restricted to propositional reasoning, autoepistemic logic has recently been generalized by Levesque [Lev87]. Modifications to autoepistemic logic were also discussed in [Kon86].

The third prominent approach is based on *minimization*. This approach is embodied in many attempts at dealing with negative knowledge in databases. If one assumes complete knowledge of the domain under consideration, one can represent negative information *implicitly*. Thus a negative fact can be assumed if its positive counterpart is not entailed by the database. The assumption of complete domain knowledge is termed the *Closed World Assumption (CWA)* by Reiter, and there has been a significant amount of research in this area. One of the most studied of the minimality-based approaches is *circumscription*, which was first proposed by McCarthy in [McC80]. The notion we will describe here is called *predicate circumscription*. This was later extended by McCarthy to *formula circumscription*, discussion of which we omit here. Predicate circumscription is tied to the notion of minimal models. Informally, the idea behind circumscribing a predicate $P$ in a formula $F$ is to produce a formula which, when conjoined with $F$, forces $P$ to be true for those atomic objects for which it is *forced* to be true by $F$, and false otherwise. This conjunction is called the *circumscription of $P$ in $F$* (written $CIRC[F; P]$). Formally, this is defined to be:

$$CIRC[F; P] =_{def} F \wedge (\forall P' \\ (F(P') \wedge (\forall x P'(x) \Rightarrow P(x))) \Rightarrow \\ (\forall x P(x) \rightarrow P'(x)))$$

where $F(P')$ is the formula $F$ with all occurrences of $P$ replaced with $P'$. Although this is a second-order formula, in many cases it reduces to a first-order formula.

The most prominent work attempting to develop a usable computational framework for default reasoning is that involving *truth maintenance systems*. Among others, this approach is exemplified by the work of Doyle [Doy79b], de Kleer [dK86], Goodwin [Goo87], and Brown [BGB87]. The work described in this paper involves propagation of uncertainty measures through a network similar to the JTMS graph described by Doyle in [Doy79b].

### 1.4 Proposed Approach

In the rest of this paper we discuss our efforts in integrating defeasible reasoning (based on nonmonotonic rules) with plausible reasoning (based on monotonic rules with partial degrees of sufficiency and necessity).

In our approach, uncertainty measures are propagated through a Doyle-JTMS graph whose labels are real-valued certainty measures. Unlike other default reasoning languages that only model the incompleteness of the information, our approach uses the presence of numerical certainty values to distinguish *quantitatively* the different admissible labelings and pick an *optimal* one.

The key idea is to exploit the information on the monotonic edges carrying uncertainty measures. A preference function, based on these measures together with nonmonotonic information, is used to select the extension that is maximally consistent with the constraints imposed by the monotonic edges. Thus, instead of minimizing the cardinality of abnormality types [McC86] or of performing temporal minimizations [Sho86], we maximize an expectation function based on the uncertainty measure. This method breaks the symmetry of the (potentially) multiple extensions in each loop by selecting a *most likely* extension. This idea is currently being implemented in PRIMO (Plausible ReasonIng MOdule), RUM's successor.

We will illustrate our approach through an example. For this purpose we will prevail upon Tweety, the much overworked flying emu. The example consists of the following default rules:

BIRD $\wedge \neg \Box$ HOPS $\rightarrow$ FLIES  
EMU $\wedge \neg \Box$ FLIES $\rightarrow$ HOPS  
FLEMU $\rightarrow$ EMU  
EMU $\rightarrow$ BIRD  
FLEMU $\rightarrow$ FLIES

The first rule states that unless it can be proven that a bird hops, assume that it flies. The second says that unless it can be proven that an emu flies assume it hops. Given that FLEMU is false, and EMU true there are two valid extensions for this set of rules. One in which FLIES is true and HOPS is false; and one in which FLIES is false and HOPS is true. As we develop this example we will show how our approach uses quantitative uncertainty measures to facilitate choice between these two valid extensions.

The following section defines PRIMO's rule-graph semantics and constraints. Section 3 describes the generation of admissible labelings (consistent extensions) and introduces an objective function to guide the selection of preferred extensions. Section 4 discusses optimization techniques (applicable on restricted classes of graphs) and heuristics (such as graph decomposition into strongly connected components), which can be used to generate acceptable approximations to the optimal solution. The conclusion section summarizes our results and defines an agenda of possible future research work.

## 2 *P*lausible *R*eason*I*ng *MO*dule

The decision procedure for a logic based on real-valued truth values may be much more computationally expensive than that for boolean-valued logic. This is because in boolean-valued logic only one proof need be found. In real-valued logic all possible proofs must be explored in order to ensure that the certainty of a proposition has been maximized.

RUM (Reasoning with Uncertainty Module), the predecessor to PRIMO, was designed as a monotonic expert system shell that handles uncertainty according to triangular norm calculi[1] It deals with the possible computational explosion by allowing only propositional acyclic[2] quantitative Horn clauses.

To avoid the computational problems associated with first order reasoning, RUM only allows propositional rules. Although the user may write first-order rules, they must be fully instantiated at run time. Thus a single written rule may give rise to many rules at run time, all of which are propositional.

RUM restricts its rules to Horn clauses; it deals with negative antecedents by treating $P$ and $\neg P$ independently. We denote *the certainty of $P$* as LB($P$). The only time $P$ and $\neg P$ will interact is when LB($P$) + LB($\neg P$) > 1 (both $P$ and $\neg P$ are believed). When this occurs a conflict handler tries to detect the source of inconsistency[3].

---

[1] Triangular norm calculi represent logical *and* as a real valued function called a t-norm, and logical *or* as a s-conorm. For an introduction to them see [Bon87]. A succinct presentation can be found in [Bon89].

[2] Unless an idempotent t-norm is used cyclic rules will cause all certainties in the cycle to converge to 0.

Due to these restrictions a simple linear time algorithm exists for propagating certainty values through RUM rules. Resolution of inconsistency by the conflict handler, however, may require cost exponential in some subset of the rules.

PRIMO (Plausible ReasonIng MOdule) is the successor to RUM designed to perform nonmonotonic reasoning. PRIMO extends RUM by allowing nonmonotonic antecedents. PRIMO also allows nonmonotonic cycles which represent conflicts between different defaults. We provide a formal overview of PRIMO below:

**Definitions:** A PRIMO specification is a triple ($L$, $I$, $J$). $L$ is a set of ground literals, such that whenever $l \in L$, $\bar{l} \in L$. For $l \in L$, LB($l$) $\in [0, 1]$ is the amount of evidence confirming the truth of $l$. $J$ is a set of justifications. Each justification is of the form:

$$\bigwedge_i ma_i \wedge \bigwedge_j nma_j \rightarrow^s c$$

where $c$ is the conclusion, $s \in [0, 1]$ is the *sufficiency* of the justification ($s$ indicates the degree of belief in the conclusion of the justification when all the antecedents are satisfied), $ma_i \in L$ are the monotonic antecedents of the justification, and $nma_j$ are the nonmonotonic antecedents of the justification. The nonmonotonic antecedents are of the form $\neg\boxed{\alpha}p$, where $p \in L$, with the semantics:

$$LB(\neg\boxed{\alpha}p) = \begin{cases} 0 & \text{if } LB(p) \geq \alpha \\ 1 & \text{if } LB(p) < \alpha \end{cases}$$

(A nonmonotonic antecedent $\neg\boxed{\alpha}p$ can be informally interpreted as "if we fail to prove proposition $p$ to a degree of at least $\alpha$.") The *input literals* $I \subset L$, are a distinguished set of ground literals for which a certainty may be provided by outside sources (e.g. user input), as well as by justifications. The certainty of all other literals can only be affected by justifications.

A PRIMO specification can also be viewed as an AND/OR graph, with justifications mapped onto AND nodes and literals mapped onto OR nodes.

**Definition:** A valid PRIMO graph is a PRIMO graph that does not contain any cycles consisting of only monotonic edges.

---
[3]Note that the above constraint on LBs implies an upperbound on LB($P$) of 1 $-$ LB($\neg P$). In the literature this is denoted as UB($P$). LB and UB are related just as support and plausibility in Dempster-Shafer, or □ and ◇ in modal logics.

**Definition:** An *admissible labeling* of a PRIMO graph is an assignment of real numbers in [0, 1] to the arcs and nodes that satisfy the following conditions:

1. the label of each arc leaving a justification equals the t-norm of the arcs entering the justification and the sufficiency of the justification and

2. the label of each literal is the s-co-norm of the labels of the arcs entering it.

A PRIMO graph may have zero, one, or many admissible labelings. An odd loop (a cycle traversing an odd number of *nonmonotonic wires*) is a necessary but not sufficient condition for a graph to have no solutions. Every even cyclic graph has at least two solutions. In these respects PRIMO is like the Doyle JTMS [Doy79a]. Proofs can be found in [Goo88].

## 2.1 PRIMO Example

We use our example to illustrate the above definitions. The default rules given in the earlier example can be turned into PRIMO justifications by adding sufficiencies to them. For example:

BIRD $\wedge \neg\boxed{.2}$ HOPS $\rightarrow^{.8}$ FLIES
EMU $\wedge \neg\boxed{.2}$ FLIES $\rightarrow^{.9}$ HOPS
FLEMU $\rightarrow^1$ EMU
EMU $\rightarrow^1$ BIRD
FLEMU $\rightarrow^1$ FLIES

In the first justification, BIRD is a monotonic antecedent and $\neg\boxed{.2}$ HOPS is a nonmonotonic antecedent. The sufficiency of the justification is .8. The first rule states that if it can be proven with certainty $\geq$ .2 that HOPS is true then the certainty of FLIES is 0; otherwise the certainty of FLIES is the t-norm of .8 and the certainty of BIRD being true.

The input literals for this example are BIRD, EMU, and FLEMU.

The PRIMO graph corresponding to the above rules is shown in Figure 1. If the user specifies that LB(BIRD) = LB(EMU) = 1, and LB(FLEMU) = 0, then there are two admissible labelings of the graph. One of these labelings is shown in Figure 2. The other can easily be obtained by changing the labeling of HOPS to 0 and FLIES to .8.

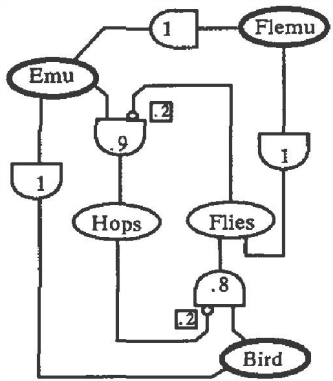

Figure 1: PRIMO Rule Graph

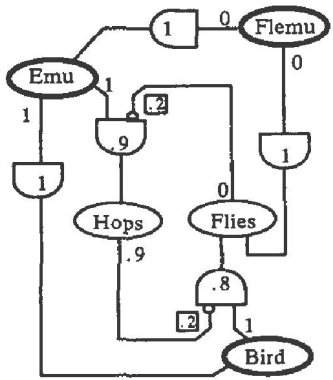

Figure 2: One Admissible Labeling

# 3 Finding Admissible Labelings

In this section we discuss an approach to propagation of constraints which is used as a preliminary step in processing a PRIMO graph before resorting to exhaustive search.

## 3.1 Propagation of Bounds

In PRIMO, propagation of bounds on LB's can be more effective than propagation of exact values alone. It may even trigger further propagation of exact values when bounds are propagated to a nonmonotonic antecedent whose value of $\alpha$ falls outside of them. Thus bounds propagation can sometimes provide an exact solution where propagation of exact values alone would not.

To propagate bounds, vertices are labeled with pairs of values representing lower and upper bounds on the exact LB of that vertex in any admissible labeling. These bounds are successively narrowed as propagation continues. For each vertex $v$ we define $LB^-(v)$ and $LB^+(v)$, the lower and upper bounds on $LB(v)$ at any given point during the computation, to be functions of the bounds then stored on the antecedents of $v$. $LB^-$ uses the lower bounds of monotonic antecedents and the upper bounds of nonmonotonic ones; $LB^+$ uses the upper bound of monotonic and the lower bound of nonmonotonic antecedents. The actual function applied to these values is the same one used to compute LB itself for that vertex. The algorithm is then:

1. Initialize every input node, where $k$ is the confidence given by the user, to $[k, 1]$ i.e. "at least $k$". Initialize every other vertex to $[0, 1]$.

2. While there exists any vertex $v$ such that the label on $v$ is not equal to $[LB^-(v), LB^+(v)]$, relabel $v$ with that value.

It can be shown that this algorithm converges in polynomial time, yields the same result regardless of the order of propagation, and never assigns bounds to a vertex which exclude any value that vertex takes on in any admissible labeling. Proofs can be found in [Goo88].

## 3.2 Example

In this section we illustrate, through the example, the bounds propagation algorithm. Figure 3 shows the labeling of the graph after the initialization step. Figure 4 shows the final bounds obtained.

The value of $LB^+$ for FLIES of .8 is derived by using $LB^-$ of HOPS (0) which gives the certainty of the premises of the justification for FLIES of 1. The t-norm of this with the sufficiency of the justification yields .8.

## 3.3 A Labeling Algorithm for PRIMO

**Definitions:** A nonmonotonic antecedent is *satisfied* if $LB^+ < \alpha$, *exceeded* if $LB^- \geq \alpha$, and *ambiguous* if $LB^- < \alpha \leq LB^+$. A labeled graph is *stable* if every vertex $v$ is labeled $[LB^-(v), LB^+(v)]$ (a graph is always stable after bounds have been propagated). In a stable graph, a *starter dependency* is an AND-vertex which has no unlabeled monotonic antecedents, no exceeded nonmonotonic antecedents, and at least one ambiguous nonmonotonic antecedent.

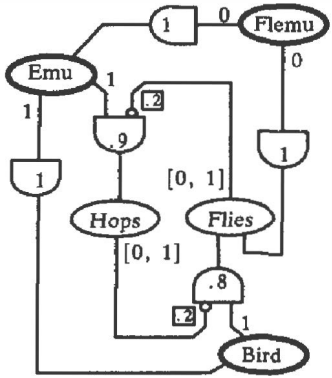

Figure 3: Initialization

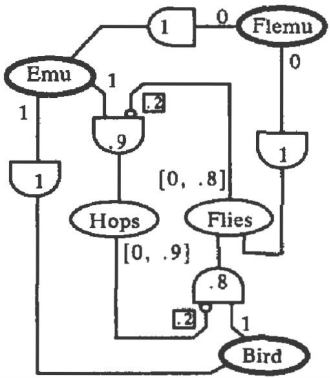

Figure 4: Final Bounds

A starter dependency must be unlabeled, with a zero $LB^-$ and a positive $LB^+$. Because PRIMO nets contain no monotonic loops, a starter dependency always exists (unless the entire graph is labeled exactly) and can be found in time linear in the size of the graph. Because the only inputs left undetermined are nonmonotonic antecedents (i.e., thresholds) a starter dependency must be labeled exactly $LB^-$ or $LB^+$ in any admissible labeling which may exist [Goo88].

One can therefore find all admissible labelings of a stable graph in time exponential in the number of starter dependencies, simply by generating each of the $2^k$ ways to label each of $k$ starter dependencies in the graph with its $LB^-$ or $LB^+$, and testing each combination for consistency.

A straightforward algorithm to do this would search the space depth-first with backtracking. Each iteration would pick a starter dependency, force it to $LB^-$ or $LB^+$, and propagate bounds again, continuing until either a solution is produced or an inconsistency is found, and then backtrack.

Inconsistencies can only occur at a starter dependency, when either (1) the starter was earlier forced to $LB^-$ (i.e., zero) and positive support for it is derived, or (2) the starter is forced to $LB^+$ (i.e., a positive value) and the last support for it becomes relabeled zero.

Practical efficiency may be greatly enhanced if the starter dependency is always chosen from a minimal strongly connected component of the unlabeled part of the graph.

Below we consider more sophisticated methods for searching this space.

### 3.4 Consistent and Preferred Extensions

The discussion and algorithm given above indicate that in a stable graph the problem of deciding upon how to resolve the ambiguous nonmonotonic wires is a boolean decision. Thus we should be able to formulate this problem in propositional logic, the satisfying assignments of which would represent the various consistent extensions of the PRIMO specification.

We now present an alternate algorithm, based on propositional satisfiability, for finding consistent extensions. We also show how this algorithm can be used to find an *optimal* extension.

In general, a set of formulae will have many extensions. Given such a set of extensions, some may be preferable to others based on the cost associated with choosing truth values for certain nodes. That is, the LB of the ambiguous antecedents will be coerced to either $LB^-$ or $LB^+$. We will prefer extensions in which the sum of the costs associated with choosing a truth value for each proposition is minimized. More formally, let $\neg\boxed{\alpha_i}\,p_i$ be the set of nonmonotonic premises from a PRIMO rule graph which are still ambiguous after the numeric bounds have been propagated; let $IC(p_i) = \frac{LB^-(p_i)+LB^+(p_i)}{2}$. $IC(p_i)$ is a measure of the current approximation of the information content in $p_i$. An optimal admissible labeling is an admissible labeling that minimizes the objective function:

$$\sum_i |IC(p_i) - FPV(p_i)| + |IC(\neg p_i) - FPV(\neg p_i)|.$$

$FPV(p_i)$, the final propositional value associated with $p_i$, will be either 0 or 1, depending on whether $p_i$ is ultimately coerced to $LB^-(p_i)$ or $LB^+(p_i)$, respectively. Thus the objective function is a measure of the distance of our current numerical approximation to the final value chosen, which we want to minimize.

Once we have made the commitment to coercing ambiguous values to either 0 or 1, solving the problem of finding extensions reduces to propositional satisfiability. Extending the problem we consider to that of *weighted satisfiability*, gives us a means of finding a preferred extension. Weighted satisfiability is defined formally below:

> Let $C$ be a weighted CNF formula, $\bigwedge_i C_i$, where each clause, $C_i = \bigvee_j p_j$, has a corresponding positive weight, $w_i$. Let $P$ be a truth assignment of the propositional variables, $p_i$, that appear in $C$. The weight of $P$ is the sum of the weights of the clauses that are made false by $P$. The weighted satisfiability problem is to find the minimum weighted truth assignment.

The optimal admissible labeling problem can be encoded as the weighted satisfiability problem in the following way:

Convert the propositional form of the given PRIMO graph into clausal form. Assign infinite weight to each of the resulting clauses. Define a function $W$ which maps literals into the interval $[0, 1]$ such that for a literal $p$,

$$W(p) = IC(p) + (1 - IC(\neg p)).$$

Next, for each ambiguous nonmonotonic premise of the form $\neg \boxed{\alpha_i} p_i$, generate two clauses:

1. $(p_i)$, with weight $W(p_i)$
2. $(\neg p_i)$ with weight $W(\neg p_i)$.

The first clause represents the cost of making $p_i$ false, the second the cost of making $\neg p_i$ false (equivalently, making $p_i$ true). A typical case is illustrated in Figure 5.

It is easy to see that the original graph has an admissible labeling if and only if there is a finitely weighted truth assignment for the corresponding instance of weighted satisfiability, and that the weighted truth assignment corresponds to minimizing the objective function given above.

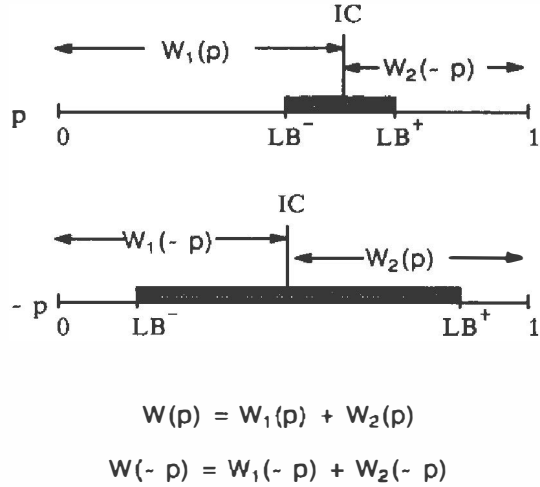

$$W(p) = W_1(p) + W_2(p)$$
$$W(\neg p) = W_1(\neg p) + W_2(\neg p)$$

Figure 5: Relationship between bounds and weights

### 3.5 Example

We now complete our example by showing how the graph in Figure 4 can be transformed into a weighted satisfiability problem which yields the optimal extension for this example. From the graph we obtained the following values:

|        | FLIES | ¬ FLIES | HOPS | ¬ HOPS |
|--------|-------|---------|------|--------|
| $LB^-$ | 0     | 0       | 0    | 0      |
| $LB^+$ | 0.8   | 0       | 0.9  | 0      |
| $IC$   | 0.4   | 0       | 0.45 | 0      |
| $W$    | 1.4   | 0.6     | 1.45 | 0.55   |

Figure 6: Evaluation of Weights

The weighted clauses, obtained from the structure of the graph and from the table above, are:

| | |
|---|---|
| FLIES ∨ HOPS | ∞ |
| ¬FLIES ∨ ¬HOPS | ∞ |
| FLIES | 1.4 |
| ¬FLIES | .6 |
| HOPS | 1.45 |
| ¬HOPS | .55 |

There are two finitely weighted truth assignments for the above set of weighted clauses. They are FLIES = True, HOPS = False with weight 2.05; and FLIES = False, HOPS = True with weight 1.95. (Remember the weight of a truth assignment is the sum of the weights of the clauses made false by the truth assignment.) Thus the optimal labeling for our example gives LB(FLIES) = 0, and LB(HOPS) = .9.

We leave it to the reader to verify that starting with LB(EMU) = .8 instead of 1 would result in the optimal labeling where LB(FLIES) = .8, LB(HOPS) = 0.

## 4 Algorithms and Heuristics

In Section 3.4 we showed how the problem we are concerned with can be posed as one of *weighted satisfiability*. Since this problem is intractable in general, we must make compromises if our system is to perform reasonably on nontrivial instances. The alternatives we consider include constraining the classes of problems we will allow (Section 4.1) or sacrificing optimality of solutions (Section 4.2).

### 4.1 Nonserial Dynamic Programming

One of the most interesting possibilities involves restricting our attention to classes of formulae which, while still intractable, have satisfiability algorithms which theoretically take much less than $O(2^n)$ time, where $n$ is the number of propositional variables. In [RH86], Hunt and Ravi describe a method based on *nonserial dynamic programming* and *planar separators* (see [BB72] and [LT80], respectively) which solves the satisfiability problem in $O(2^{\sqrt{n}})$ time for a subclass of propositional clauses that can be mapped in a natural way to planar graphs[4]. In [Fer88] Fernandez-Baca discusses an alternative construction for planar satisfiability and an extension to *weighted satisfiability*. He also presents a similar algorithm for another interesting class of problems, where the graph corresponding to the set of clauses has bounded *bandwidth*. Hunt [Hunt] has shown that similar results hold for a large class of problems which have graphs with *bounded channel width*. Each of these is in some sense a measure of the complexity of the clausal form of the problem. If they are much smaller than the number of variables in the problem, weighted satisfiability can be solved relatively quickly for large instances.

### 4.2 Heuristics

Depending on the size of the graph and the deadline imposed on the system by the outside world, time to find an optimal extension may not be available. Under these circumstances, we need to use a heuristic that, without guaranteeing an optimal solution, will find a satisficing solution while exhibiting reasonable performance characteristics.[5]

The following heuristics can be applied to the PRIMO graph, after the propagation of bounds, or to the problem encoded in terms of weighted satisfiability.

As initial conditions we assume a set of nodes **P**, which is a subset of the original set of nodes in the graph. Each element of **P** has an associated pair of lower and upper bounds. We sort the elements in **P** such that $| IC(p_i) - 0.5 | \geq | IC(p_{i+1}) - 0.5 |$. By sorting the elements in **P** based on decreasing information content, we are trying to first coerce the labeling of those nodes for which we have the strongest constraints.

We can now use a variety of search strategies, such as the iteratively deepening hill-climbing search, or beam-search to (locally) minimize the objective function defined in Section 3.4, subjected to the *consistency constraints* dictated by the graph topology.

### 4.3 Strongly Connected Components

Thus far we have presented our algorithms as if they were to work on the entire PRIMO rule graph. Even the heuristic presented would bog down on rule graphs of realistic size.

As a result, several optimizations are essential in practice, even though they do not affect the theoretical worst case complexity. The entire initial graph can be decomposed into strongly connected components (SCCs), which are attacked one at a time (using whatever algorithm or heuristic is deemed appropriate) "bottom up".

This idea was first used for JTMSs in [Goo87]. As in the JTMS, there is no guarantee that one can avoid backtracking: a low level SCC may have several solutions, and a higher SCC dependent upon it may become unsolvable if the wrong choice is made lower down. However, this strategy seems likely to be helpful in practice.

---

[4]It is shown in [Lic82] that the satisfiability problem for this class is NP-complete [GJ79]. Thus the existence of a polynomial time decision procedure is highly unlikely.

[5]As any other heuristic, there is no guarantee that its worst case performance can improve that of an exhaustive search.

## 4.4 Compile Time Options

Decomposing the initial graph into SCCs is one form of preprocessing that can be done at compile time in an attempt to facilitate faster run time processing. Much more processing could be done at compile time, at potentially great savings at run time. This section briefly summarizes various options we are exploring for dividing the task of generating optimal (good) admissible labelings between run time and compile time components.

- Precompute all admissible labelings at compile time. At run time eliminate those labelings that are not currently valid, and choose the optimal labeling from those remaining.

- Precompute a subset of the admissible labelings at compile time. At run time eliminate those labelings that are not currently valid, and choose the optimal labeling from those remaining. If all the precomputed labelings have been eliminated, additional labelings must be generated.

- Precompute one default admissible labeling optimized according to static uncertainty and utility information. If this labeling is no longer valid at run time, additional labelings must be generated.

- Precompile the graph into some canonical form that will allow easier generation of admissible labelings at run time. Two possible such forms are prime implicants (similar to the ATMS [RdK87]) and Gröbner bases [KN85]. At run time the compiled form can be used to generate an admissible labeling.

Although all the above options may result in great savings at run time, many of them share the problem of potentially incurring exponential time and space costs in the worst case. At present, we are beginning to experiment with these options in an attempt to determine which will work best in practice.

## 5 Conclusions

We have presented an approach that integrates nonmonotonic reasoning with the use of quantitative information as a criterion for model preference. This represents a major departure from existing paradigms, which normally fail to account for one or the other. We have also identified several methods for coping with the inherent intractability involved in such reasoning. We feel that this is a promising approach, but this work is at a preliminary stage. As a result, there are a number of questions which we are considering now. We list some of them below.

- We have previously noted that there are some correspondences between the PRIMO rule graph and that of the JTMS. Their exact relationship (if indeed one exists) is not well understood and needs to be explored.

- The dynamic programming algorithms discussed in Section 4.1 may help us to deal with large problem instances under certain structural constraints on the allowed propositional formulae. The results discussed, however, are based on asymptotic bounds. We have begun to implement these algorithms, but do not know at this point whether they will perform satisfactorily in practice. We also need to determine how well the heuristics we have described will perform.